\documentclass{fairmeta}
\usepackage{amsmath}
\usepackage{enumerate} 
\usepackage{bm}
\usepackage{floatflt}
\usepackage{algpseudocode}
\usepackage{amsfonts}
\usepackage{amsthm}
\usepackage{newtxtt}
\usepackage{colortbl} 
\usepackage{cleveref}
\usepackage{diagbox} 
\usepackage[utf8]{inputenc}
\usepackage{textgreek}
\usepackage{colortbl}
\usepackage{nicematrix}
\usepackage{makecell}
\usepackage{float}
\usepackage{arydshln}
\usepackage[frozencache,cachedir=.]{minted}
\usepackage{caption}
\usepackage{subcaption}
\usepackage{tcolorbox}
\usepackage{amssymb}
\usepackage{xspace}
\usepackage{wrapfig}
\usepackage{adjustbox}
\usepackage{tabularx}
\usepackage{booktabs}
\usepackage{mathtools}
\usepackage{wrapfig}
\usepackage{amssymb}
\usepackage{graphicx}

\usepackage{silence}
\makeatletter
\patchcmd{\wrong@fontshape}{\@gobbletwo}{}{}{}
\makeatother
\definecolor{upColor}{RGB}{17,138,21}
\definecolor{downColor}{RGB}{174,36,67}

\newtheorem{theorem}{Theorem}[]

\newtheorem{remark1}[theorem]{Remark}

\usepackage[linesnumbered,ruled,vlined]{algorithm2e}
\newcommand{\KwParam}{\textbf{Parameters:} }
\usepackage{graphicx}
\graphicspath{{./fig/}}

\usepackage[utf8]{inputenc}

\usepackage{booktabs}   
\usepackage{multirow}   
\usepackage{xcolor}     
\usepackage{graphicx}   
\usepackage{pifont}     
\usepackage{array}      
\usepackage{float}
\usepackage{setspace}

\newcommand{\best}[1]{\textbf{#1}}
\newcommand{\second}[1]{\underline{#1}}
\newcolumntype{x}[1]{>{\centering\arraybackslash}p{#1}}

\title{SSVP: Synergistic Semantic-Visual Prompting for Industrial Zero-Shot Anomaly Detection}

\author[1,2]{Chenhao Fu}
\author[2]{Han Fang}
\author[2]{Xiuzheng Zheng}
\author[1]{Wenbo Wei}
\author[1]{Yonghua Li}
\author[2]{Hao Sun}
\author[2]{Xuelong Li}
\affiliation[1]{Beijing University of Posts and Telecommunications}

\affiliation[2]{Institute of Artificial Intelligence (TeleAI), China Telecom}

\metadata[Correspondence to]{Hao Sun (\email{sun.010@163.com}), Xuelong Li (\email{xuelong\_li@ieee.org})}

\begin{document}

\abstract{
Zero-Shot Anomaly Detection (ZSAD) leverages Vision-Language Models (VLMs) to enable supervision-free industrial inspection. However, existing ZSAD paradigms are constrained by single visual backbones, which struggle to balance global semantic generalization with fine-grained structural discriminability. To bridge this gap, we propose Synergistic Semantic-Visual Prompting (SSVP), that efficiently fuses diverse visual encodings to elevate model's fine-grained perception. Specifically, SSVP introduces the Hierarchical Semantic-Visual Synergy (HSVS) mechanism, which deeply integrates DINOv3's multi-scale structural priors into the CLIP semantic space. Subsequently, the Vision-Conditioned Prompt Generator (VCPG) employs cross-modal attention to guide dynamic prompt generation, enabling linguistic queries to precisely anchor to specific anomaly patterns. Furthermore, to address the discrepancy between global scoring and local evidence, the Visual-Text Anomaly Mapper (VTAM) establishes a dual-gated calibration paradigm. Extensive evaluations on seven industrial benchmarks validate the robustness of our method; SSVP achieves state-of-the-art performance with 93.0\% Image-AUROC and 92.2\% Pixel-AUROC on MVTec-AD, significantly outperforming existing zero-shot approaches.

}



\maketitle

\section{Introduction}

Industrial Anomaly Detection (IAD) is vital for smart manufacturing and industrial quality inspection, ensuring production reliability~\citep{ref_2024survey, ref_pr_efficient, ref_pr_multiad, ref_pr_slsg, ref_pr_multiexp}. Yet, traditional supervised models struggle with the inherent data scarcity and frequent category shifts~\citep{ref_pr_oneshot, ref_pr_dcad}. Zero-Shot Anomaly Detection (ZSAD) overcomes this by eliminating the need for target domain labels~\citep{ref_transfer}, aiming to identify unseen defects via generalized knowledge transfer. Recently, CLIP-based VLMs~\citep{ref_pr_vlm, ref_pr_vlm2} have enabled language-guided visual detection by constructing a unified semantic space. While early methods used static templates~\citep{ref_winclip, ref_aprilgan}, current research prioritizes Dynamic Prompting~\citep{ref_cocoop, ref_anomalyclip, ref_adaclip}. By adapting to specific inputs, this approach handles complex defects more effectively than fixed descriptions.

\begin{figure}[h]
    \centering
    \includegraphics[width=1\linewidth, trim=15mm 170mm 60mm 15mm, clip]{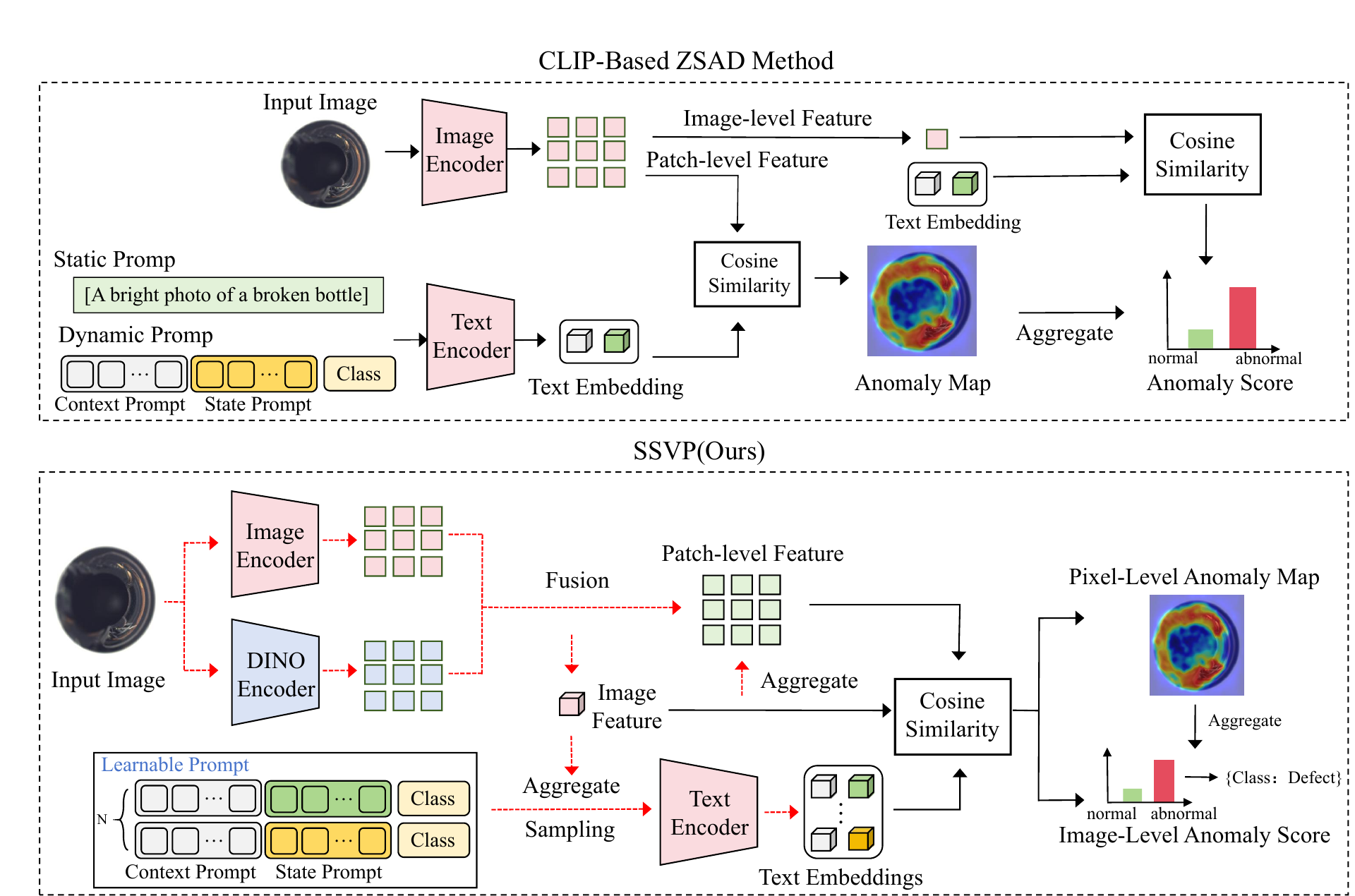}
    \caption{Conceptual comparison of ZSAD paradigms: unlike traditional shallow fusion methods, SSVP establishes a deep synergistic flow for adaptive prompt generation. The red dashed lines highlight the distinction between our SSVP method and traditional methods.}
    \label{fig:framework_compare}
\end{figure}

Despite these advancements, existing dynamic prompting methods face three systemic limitations in complex scenarios. First, VLM's global alignment compromises fine-grained discriminability, failing to capture minute distortions essential for precise industrial inspection. Second, current probabilistic prompting~\citep{ref_bayes_pfl} lacks deep cross-modal interaction. By relying on shallow linear superposition to integrate visual priors, these methods fail to effectively enable text embeddings to adaptively anchor to specific anomaly patterns. Third, global scoring is misaligned with local evidence. The heavy reliance on global cosine similarity often causes subtle anomaly signals to be overshadowed by dominant background noise, leading to missed detections in high-precision tasks. To intuitively illustrate these limitations, we present a conceptual comparison between existing paradigms and our proposed framework in Figure \ref{fig:framework_compare}.
To address these challenges, we propose the Synergistic Semantic-Visual Prompting (SSVP) framework. Unlike previous approaches that treat modalities in isolation, SSVP integrates a deep synergistic flow that harmonizes CLIP's generalized semantic reasoning with DINO's fine-grained structural localization.

Specifically, we instantiate this method through three technically distinct modules. 
First, to bridge the feature granularity gap, we design the Hierarchical Semantic-Visual Synergy (HSVS) mechanism. In detail, HSVS employs an Adaptive Token Features Fusion (ATF) block that utilizes dual-path cross-modal attention to align diverse features, explicitly injecting DINOv3's multi-scale structural priors into CLIP's semantic manifold via learnable projection matrices.
Second, to alleviate the limitations of deterministic prompting, we propose the Vision-Conditioned Prompt Generator (VCPG). This module leverages a Variational Autoencoder (VAE) to model the latent distribution of visual anomalies and employs a text-latent cross-attention mechanism. This design allows text embeddings to dynamically "retrieve" and integrate generative visual biases via the reparameterization trick, ensuring robust anchoring to unseen defect patterns.
Third, to overcome the global-local disconnection, we construct the Visual-Text Anomaly Mapper (VTAM). This module implements an Anomaly Mixture-of-Experts (AnomalyMoE) paradigm driven by a dual-gating mechanism, encompassing global scale selection and local spatial filtering. By adaptively routing multi-scale features and calibrating global scores with prominent local evidence, VTAM ensures precise anomaly localization.

Overall, the core contributions of this study are summarized as follows:
    
    
\textbf{Hierarchical Semantic-Visual Synergy:} HSVS boosts fine-grained perception via deep fusion of multi-scale structural priors.

\textbf{Visual Conditional Prompting:} VCPG uses generative visual priors to modulate text embeddings, enhancing prompt diversity and semantic consistency.
    
\textbf{Locally Enhanced Discrimination:} VTAM calibrates global scores via an Anomaly Mixture-of-Experts strategy to resolve insensitivity to subtle defects.

\section{Related Work}
\label{sec:related_work}

\subsection{Zero-Shot Anomaly Detection}
The integration of Vision-Language Models (VLMs) has established a flexible paradigm for Zero-Shot Anomaly Detection (ZSAD)~\citep{ref_zsad1, ref_zsad2}. By aligning visual data with linguistic concepts, these models enable the recognition of anomalies through descriptions rather than examples. This approach differs fundamentally from traditional supervised strategies. Traditional methods rely heavily on extensive training samples from specific inspection scenarios. In contrast, ZSAD prioritizes cross-domain generalization. It effectively leverages universal knowledge learned from large-scale source datasets. Consequently, the system can detect anomalies in unseen target domains without requiring additional training on them.
Pioneering works have utilized CLIP \citep{ref_2024survey, ref_clip} through various strategies to adapt to this task. For instance, some approaches employ multi-scale sliding windows to improve localization \citep{ref_winclip}, while others use linear embedding adapters \citep{ref_aprilgan} or prototype memory banks to estimate rarity \citep{ref_rareclip}. While effective for object-level classification, these methods predominantly rely on coarse feature matching, which often limits their capability to capture fine-grained textural and structural defects due to the lack of dense visual-semantic interaction.

\subsection{Prompt Learning methods}
Prompt engineering is pivotal for adapting VLMs to anomaly detection, evolving from static templates to probabilistic generation\citep{ref_vlm1, ref_vlm2}. Early methods \citep{ref_winclip, ref_aprilgan} relied on manually designed static prompts, which struggle to cover diverse defect types. To improve adaptation, dynamic prompting methods \citep{ref_cocoop, ref_anomalyclip, ref_adaclip} were introduced, optimizing learnable context vectors to achieve instance-specific semantic adjustment. However, these methods remain deterministic, converging to point estimates that may not fully encompass the long-tail distribution of unseen anomalies. Consequently, recent research has pivoted towards probabilistic modeling \citep{ref_bayesian_prompt}. Notably, Bayes-PFL \citep{ref_bayes_pfl} utilizes Variational Autoencoder \citep{ref_vae} to generate diverse prompt distributions via latent sampling. Despite this progress, current generative methods typically condition generation solely on simple CLIP text embeddings, thereby missing the fine-grained local cues necessary for precise defect localization.

\subsection{Visual Feature Representation}
The efficacy of anomaly detection depends on the granularity of visual representations. CLIP \citep{ref_clip, ref_pr_msgclip} leverages massive image-text pairs to establish a robust alignment between visual features and linguistic concepts. This grants it robust global semantic understanding. However, it lacks the pixel-level discriminability required for industrial inspection. In contrast, DINO \citep{ref_dinov2, ref_dinov3} leverages large-scale self-supervised training to capture dense structural and textural details. Consequently, DINO offers fine-grained perception that effectively complements CLIP's semantic capabilities.
Approaches in the unimodal AD landscape, employing techniques like memory banks \citep{ref_patchcore, ref_anomalydino, ref_efficientad, ref_foundation} or normalizing flows \citep{ref_simplenet, ref_fastflow}, have proven highly effective. Yet, a major limitation is their lack of zero-shot inference capabilities.
A primary goal in ZSAD is to enhance generalization capabilities. To achieve this, it is beneficial to combine multiple models and utilize their complementary strengths. However, integrating these models effectively remains a significant challenge. While recent works like FiLo \citep{ref_filo} and MuSc \citep{ref_musc} explore multi-scale alignment, simple fusion strategies often suffer from feature conflict due to distinct manifold topologies. This underscores the necessity for a deep synergy mechanism, as proposed in SSVP, to effectively inject discriminative structural priors into the semantic space.

\subsection{Feature Synergy Enhancement Mechanisms}
\label{sec:related:synergy}
Synergistic methods that integrate multi-modal and multi-scale features have gained prominence across diverse computer vision domains, ranging from medical imaging~\citep{ref_syn1} and industrial fault diagnosis~\citep{ref_syn2} to multi-modal sentiment analysis~\citep{ref_pr_mdcm}, autonomous driving~\citep{ref_syn5}, and general visual recognition~\citep{ref_syn3,ref_syn4}. These achievements underscore the potential of collaborative intelligence. Notably, AI Flow~\citep{ref_aiflow} introduces the concept of interaction-based emergence. It posits that the dynamic coordination of diverse agents can transcend the limitations of individual components. Inspired by AI Flow, SSVP adopts this concept by establishing a robust synergy between semantic reasoning and visual perception. Traditional methods typically rely on the static concatenation of features. However, this approach limits the effective interaction between different modalities. To address this, our method introduces a dynamic fusion process. Specifically, fine-grained visual details actively inject spatial information into global semantic representations. This process iteratively calibrates high-level understanding with local texture details. Consequently, our method outperforms standard strategies in detecting subtle defects.

\section{Method}
\label{sec:method}

\begin{figure*}[t]
\centering
\includegraphics[width=1.0\textwidth, trim=40mm 120mm 20mm 90mm, clip]{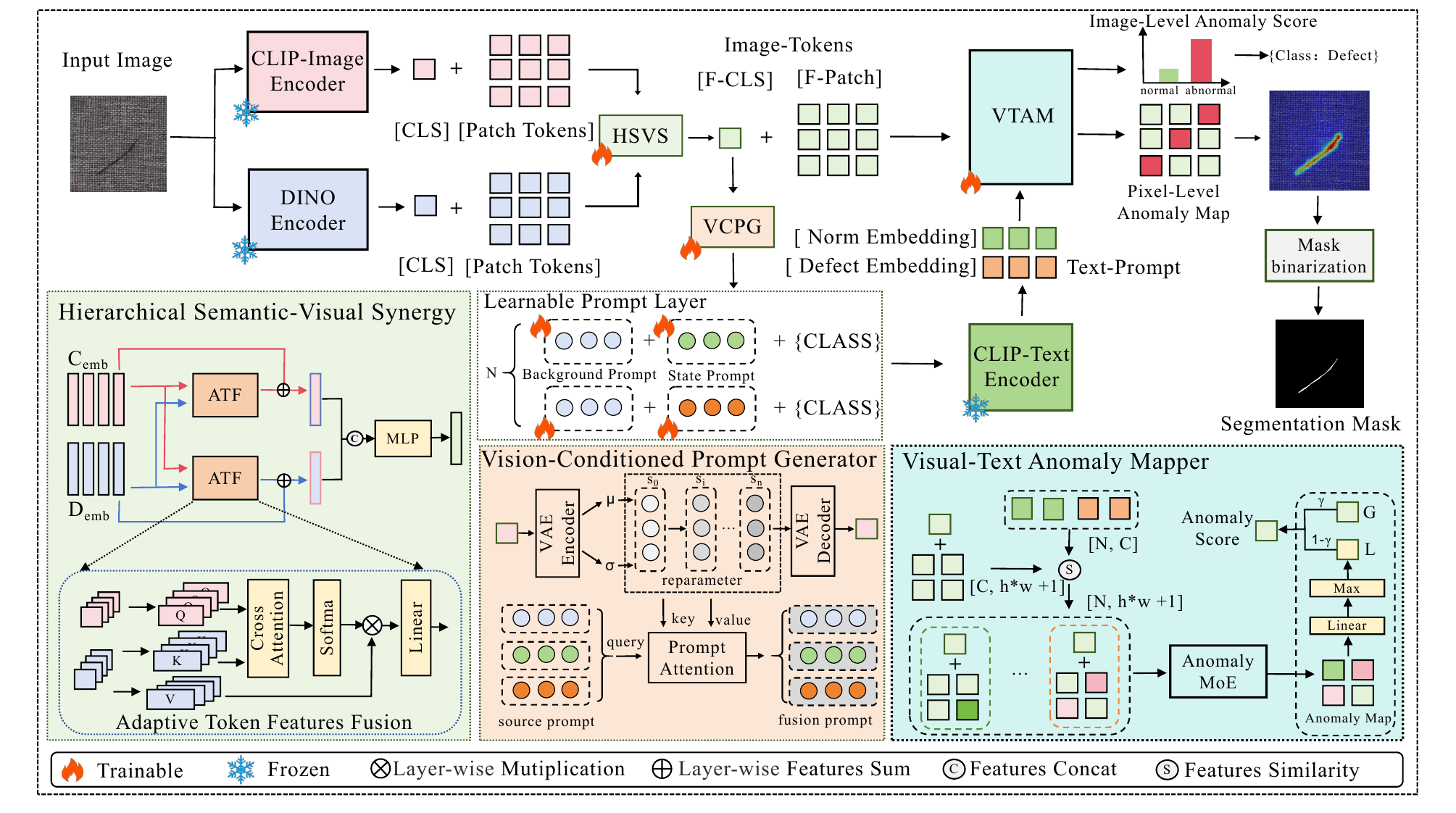}  
\caption{Framework of Synergistic Semantic-Visual Prompting (SSVP). Three core modules: HSVS aligns semantic and structural features; VCPG generates vision-conditioned prompts; VTAM refines scores via local-global interaction.}
\label{fig:framework}

\end{figure*}

In this section, we elaborate on the proposed SSVP framework. As illustrated in Figure \ref{fig:framework}, SSVP is designed to resolve the critical challenges of feature granularity bottlenecks, semantic drift, and coarse discrimination in zero-shot anomaly detection through deep cross-modal synergy. Through a synergistic fusion of high-level semantic contexts and multi-scale structural details, the framework establishes a robust visual-semantic alignment mechanism capable of identifying unseen defects without target-domain supervision.
The framework consists of three logically coupled modules. (1) Hierarchical Semantic-Visual Synergy (HSVS) constructs robust visual representations by aligning and injecting DINO fine-grained structural features into the CLIP semantic manifold. (2) Vision-Conditioned Prompt Generator (VCPG) injects enhanced visual features as strong priors into the latent space and utilizes cross-modal attention to guide text embeddings to precisely anchor on defect regions. (3) Visual-Text Anomaly Mapper (VTAM) establishes a local-global interaction paradigm to dynamically rectify global scores with local details for enhanced detection sensitivity. The detailed forward inference pipeline is presented in Algorithm \ref{alg:ssvp_forward}. 

\begin{algorithm}[t]
\setstretch{0.9} 

\caption{Forward Inference Pipeline of SSVP}
\label{alg:ssvp_forward}
\SetAlgoLined
\KwIn{Image $I$, Visual Encoders $\mathcal{E}_{clip}, \mathcal{E}_{dino}$, Text Encoder $\mathcal{T}_{clip}$}
\KwParam{Learnable Parameters $\Theta = \{\Theta_{HSVS}, \phi_{VAE}, \Theta_{VCPG}, \Theta_{MoE}, \mathcal{P}_n, \mathcal{P}_a, \alpha\}$}\;
\KwOut{Anomaly Map $\mathcal{P}_{map}$, Anomaly Score $S_{final}$, Intermediate Vars $\Phi_{aux}$}

\tcp{Feature Extraction}
$\{F^c_g, F^c_l\} \leftarrow \mathcal{E}_{clip}(I)$, $\{F^d_g, F^d_l\} \leftarrow \mathcal{E}_{dino}(I)$\;

\tcp{Stage 1: Hierarchical Semantic-Visual Synergy (HSVS)}
Project to subspace: $Q_{(\cdot)}, K_{(\cdot)}, V_{(\cdot)} \leftarrow \text{Proj}(F^c_l, F^d_l; \Theta_{HSVS})$\;
Bi-Attn: $\mathbf{A}_{c \leftrightarrow d} \leftarrow \text{Bi-Attn}(Q, K, V)$\;
Synergistic Features: 
$V_{syn}^{global}, V_{syn}^{local} \leftarrow F^c + \text{MLP}([\mathbf{A}_{c \to d}\|\mathbf{A}_{d \to c}])$ \tcp*[r]{Eq.~\ref{eq:hsvs_local}}

\tcp{Stage 2: Vision-Conditioned Prompt Generator (VCPG)}
Init embeddings: $T_{init} \leftarrow \mathcal{T}_{clip}(\mathcal{P}_n \cup \mathcal{P}_a)$\;
VAE Inference: $\mu, \sigma \leftarrow E_{\phi}(V_{syn}^{global}; \phi)$; \ $z \leftarrow \mu + \epsilon \sigma$ \tcp*[r]{Reparameterization}
Visual Injection: $\delta_{inj} \leftarrow \text{CrossAttn}(T_{init}, z; \Theta_{VCPG})$; \ $T_{final} \leftarrow T_{init} + \alpha \delta_{inj}$\;

\tcp{Stage 3: Anomaly Mixture-of-Experts (AnomalyMoE)}
$\mathcal{P}_{map} \leftarrow \text{AnomalyMoE}(\{V_{syn}^{local}\}, V_{syn}^{global}, T_{final}; \Theta_{MoE})$ \tcp*[r]{Algorithm~\ref{alg:anomaly_moe}}

\tcp{Final Scoring}
Local Evidence: $S_{local} \leftarrow \text{Mean}(\text{SpatialMaxPool}(\mathcal{P}_{map}))$\;
Global Prior: $S_{global} \leftarrow \text{CosineSim}(V_{syn}^{global}, T_{final})$\;
Final Score: $S_{final} \leftarrow (1-\gamma) S_{global} + \gamma S_{local}$ \tcp*[r]{Eq.~\ref{eq:final_score}}

Store Aux Vars: $\Phi_{aux} \leftarrow \{z, V_{syn}^{global}, T_{init}, T_{final}\}$\;
\Return $\mathcal{P}_{map}, S_{final}, \Phi_{aux}$
\end{algorithm}

\subsection{Hierarchical Semantic-Visual Synergy (HSVS)}
\label{subsec:hsvs}

To bridge the gap between the global semantics of CLIP and the fine-grained structures of DINOv3, we propose the HSVS mechanism. The core component of HSVS is the Adaptive Token Features Fusion (ATF) module. ATF constructs a refined feature interaction flow based on dual-path cross-modal attention. First, given an input image $I$, we extract features using CLIP and DINOv3 encoders and decouple them into global class tokens and local patch feature sequences:
\begin{equation}
    \{ F_{g}^{c}, F_{l}^{c} \} = \mathcal{E}_{clip}(I), \quad \{ F_{g}^{d}, F_{l}^{d} \} = \mathcal{E}_{dino}(I) ,%
\end{equation}
where $F_{l}^{c} \in \mathbb{R}^{N \times D_c}$ and $F_{l}^{d} \in \mathbb{R}^{N \times D_d}$ denote the local features of CLIP and DINO. ATF processes the global and local pairs independently. Using the local branch for illustration, to align these distinct features within a shared latent subspace, we first employ learnable projection matrices to transform the features into Query, Key, and Value embeddings. Then, we construct two parallel interaction paths:
\begin{equation}
    \begin{aligned}
        Q_c &= F_{l}^{c} W_{Q}^{c}, \quad K_d = F_{l}^{d} W_{K}^{d}, \quad V_d = F_{l}^{d} W_{V}^{d},  \\
        Q_d &= F_{l}^{d} W_{Q}^{d}, \quad K_c = F_{l}^{c} W_{K}^{c}, \quad V_c = F_{l}^{c} W_{V}^{c} ,%
    \end{aligned}
\end{equation}
where $W_{(\cdot)}^{c} \in \mathbb{R}^{D_c \times d_{head}}$ and $W_{(\cdot)}^{d} \in \mathbb{R}^{D_d \times d_{head}}$ denote the projection weights for the respective feature paths. Subsequently, we compute dual attention maps to integrate complementary information. The first path queries DINO features with CLIP semantics, while the second path performs the reverse operation. The normalized cross-modal attention weights are calculated as:
\begin{equation}
    \begin{aligned}
        \mathbf{Attn}_{c \to d} &= \text{Softmax}\left( \frac{Q_c (K_d)^\top}{\sqrt{d_{head}}} \right) V_d ,\\
        \mathbf{Attn}_{d \to c} &= \text{Softmax}\left( \frac{Q_d (K_c)^\top}{\sqrt{d_{head}}} \right) V_c ,%
    \end{aligned}
\end{equation}
where $\mathbf{Attn}_{c \to d}$ explicitly injects fine-grained geometric priors into the semantic manifold, while $\mathbf{Attn}_{d \to c}$ provides reciprocal structural reinforcement. After obtaining the dual-path context features, we apply Layer Normalization (LN) and concatenate them along the channel dimension to form a joint descriptor $Z_{joint}$:
\vspace{-0.5cm} 
\begin{equation}
    Z_{joint} = \left[ \text{LN}(\mathbf{Attn}_{c \to d}) \mathbin{\|} \text{LN}(\mathbf{Attn}_{d \to c}) \right] ,%
\end{equation} 
where $\mathbin{\|}$ denotes the concatenation operation. To achieve deep feature fusion and introduce non-linearity, $Z_{joint}$ is fed into a two-layer Multilayer Perceptron (MLP) activated by GELU. Finally, to maintain the distributional stability of the original semantic space, we employ a residual connection to add the fused features back to the original representation, yielding the final local synergistic feature $V_{syn}^{local}$:
\begin{equation}
    V_{syn}^{local} = F_{l}^{c} + \mathcal{F}_{mlp}(Z_{joint}) .%
    \label{eq:hsvs_local}
\end{equation}
Similarly, the global synergistic feature $V_{syn}^{global}$ is computed via the same ATF logic. The final output set $\{V_{syn}^{global}, V_{syn}^{local}\}$ thus combines the dual advantages of zero-shot semantic understanding and pixel-level structural awareness.

\subsection{Vision-Conditioned Prompt Generator (VCPG)}
\label{subsec:vcpg}

The core innovation of SSVP lies in the VCPG module. Traditional dynamic prompting is often limited to static representations, failing to capture the long-tail distribution of unseen anomalies. VCPG resolves this by introducing a ``visual-latent'' interaction mechanism, utilizing variational inference to transform fine-grained visual prompts into additive semantic biases.

\subsubsection{Structured Learnable Prompt Embeddings}
To distinguish between invariant environmental context and specific anomaly states, we propose a Decoupled Prompt Construction Strategy. We simplify the prompt parameterization by decomposing it into shared and state-specific components. Specifically, the general prompt template $t$ is defined as:
\begin{equation}
    t = [V]_{bg}, [V]_{state}, [CLASS] .%
\end{equation}
In this formulation:
$[V]_{bg} = \{v_1, \dots, v_L\}$ denotes the parameter-shared background vector sequence. Regardless of the sample category (normal or abnormal), this component aims to learn domain-invariant environmental semantics (e.g., texture styles, lighting conditions).
$[V]_{state}$ denotes the state-specific vector sequence. We learn a set of $[V]_{normal}$ for the normal prompt bank $\mathcal{P}_n$, and a separate set of $[V]_{abnormal}$ for the abnormal prompt bank $\mathcal{P}_a$.
This design forces the model to focus on state discrepancies rather than background variations. These sequences are processed by the text encoder $\mathcal{T}$ to generate the initial text embedding $T_{init}$.

\subsubsection{Variational Visual Modeling and Attention Interaction}
Inspired by Bayes-PFL \citep{ref_bayes_pfl}, we introduce a visual latent injection mechanism. HSVS features are encoded via a VAE into latent representations, which interact with learnable prompts through cross attention. This enables text semantics to adapt dynamically to visual anomaly patterns.
To capture the uncertainty of anomaly appearances, we employ a VAE to learn the latent manifold. We utilize the global synergistic feature $V_{syn}^{global}$ derived from the HSVS as the input observation. The probabilistic encoder $E_\phi$, composed of two parallel MLPs $\mathcal{F}_\mu$ and $\mathcal{F}_\sigma$, maps the visual feature to the parameters of the latent distribution:
\begin{equation}
    \mu = \mathcal{F}_\mu(V_{syn}^{global}), \quad \log\sigma^2 = \mathcal{F}_\sigma(V_{syn}^{global}) ,%
\end{equation}
\vspace{-0.5cm} 

where $\mu$ and $\sigma^2$ represent the mean and variance of the latent Gaussian distribution, respectively.
To ensure differentiability during back-propagation, we apply the Reparameterization Trick. We sample a standard Gaussian noise $\epsilon \sim \mathcal{N}(0, \mathbf{I})$ and generate the latent variable $z$ via an affine transformation:
\begin{equation}
    z = \mu + \epsilon \odot \exp\left(\frac{\log\sigma^2}{2} \right) .%
\end{equation}
Here, $z$ denotes the Visual Latent Bias, which encodes the global anomaly distribution of the input image. To regularize the latent space toward a standard normal distribution, we optimize the VAE by maximizing the Evidence Lower Bound (ELBO). This method aligns with established reconstruction-based paradigms in anomaly detection \citep{ref_reverse_distillation}. The VAE loss $\mathcal{L}_{VAE}$ is formulated as:
\begin{equation}
    \mathcal{L}_{VAE} = \|V_{syn}^{global} - D_\theta(z)\|_2^2 + \beta \cdot D_{KL}(q_\phi(z|V_{syn}^{global}) \| \mathcal{N}(0, \mathbf{I})) ,%
    \label{eq:vae_loss}
\end{equation}
where $D_\theta$ denotes the decoder and $\beta$ serves as a balancing coefficient. During inference, we employ a Text-Latent Cross-Attention module to modulate text semantics using the sampled $z$. Specifically, this mechanism treats the initial text embedding $T_{init}$ as the Query and maps the latent bias $z$ to Key and Value embeddings to generate the visually enriched residual $\delta_{inj}$:
\begin{equation}
    \begin{aligned}
        Q_{text} &= T_{init}W_Q, \quad K_{z} = zW_K, \quad V_{z} = zW_V, \\
        \delta_{inj} &= \text{Softmax}\left( \frac{Q_{text} (K_{z})^\top}{\sqrt{d_{k}}} \right) V_{z} .%
    \end{aligned}
\end{equation}

\subsubsection{Gated Injection with Margin-Based Constraint}

Finally, we inject the generated visual bias into the text embeddings. To prevent excessive visual information from distorting the original class distribution, we introduce an Adaptive Gating Mechanism alongside a Margin-based Semantic Regularization.
We inject the computed visual bias $\delta_{inj}$ into the initial text embedding through a learnable gating scalar $\alpha$ (initialized to 0):
\begin{equation}
    T_{final} = \text{LayerNorm}(T_{init} + \alpha \cdot \delta_{inj}) .%
\end{equation}
The gating scalar $\alpha$ adaptively regulates the strength of the injected visual bias based on the uncertainty of the input image. To balance semantic consistency with visual specificity, we avoid rigidly constraining $T_{final}$ to the initial anchor. We impose a relaxed constraint, formulating the regularization loss $\mathcal{L}_{reg}$ as:
\begin{equation}
    \mathcal{L}_{reg} = \max\left(0, \xi - \frac{T_{final} \cdot \text{sg}(T_{init})^\top}{\|T_{final}\| \|\text{sg}(T_{init})\|}\right) ,%
    \label{eq:reg_loss}
\end{equation}
where $\xi$ represents a predefined similarity threshold, and $\text{sg}(\cdot)$ denotes the stop-gradient operator. This formulation implies that penalties are incurred only when the cosine similarity drops below $\xi$. Effectively, this mechanism defines a semantic neighborhood: as long as $T_{final}$ remains within this region, the model is permitted to leverage visual bias to optimize anomaly detection performance, thereby preventing the representation from collapsing into a static embedding.

\subsection{Visual-Text Anomaly Mapper (VTAM)}
\label{subsec:vtam}

Existing ZSAD methods are limited by misaligned global and local representations. Global semantics often ignore fine-grained anomaly signals, while static aggregation fails to adapt to defects manifesting at varying scales. To address this, we introduce an Anomaly Mixture-of-Experts (AnomalyMoE) module that dynamically routes multi-scale features via a dual-gating mechanism, ensuring precise localization prior to global calibration.

\subsubsection{Pixel-Wise Anomaly Probability Map Construction}
To initiate the localization process, we first compute layer-wise anomaly probability maps using the synergistic features derived from the HSVS module and the optimized anomaly prompts. Given the synergistic local feature map $V_{syn}^{local, (l)} \in \mathbb{R}^{H_l \times W_l \times D}$ at the $l$-th layer and the visually-conditioned prompt $T_{final}$, we calculate the dense cosine similarity to generate the raw anomaly map $\mathcal{M}^{(l)}$:
\begin{equation}
    \mathcal{M}^{(l)}_{h,w} = \frac{V_{syn}^{local, (l)}(h,w) \cdot (T_{final})^\top}{\|V_{syn}^{local, (l)}(h,w)\| \|T_{final}\|} .%
    \label{eq:raw_map}
\end{equation}
To ensure numerical stability and normalize scales across different layers, we transform these raw similarity scores into pixel-wise anomaly probabilities via a Softmax operation applied across the normal and abnormal channels:
\begin{equation}
    \mathcal{P}^{(l)}_{anom}(h,w) = \frac{\exp(\mathcal{M}^{(l)}_{anom}[h,w] / \tau)}{\exp(\mathcal{M}^{(l)}_{norm}[h,w] / \tau) + \exp(\mathcal{M}^{(l)}_{anom}[h,w] / \tau)}
    \label{eq:prob_map} .%
\end{equation}
Here, $\mathcal{P}^{(l)}_{anom} \in [0,1]^{H \times W}$ represents the initial confidence of the $l$-th expert. $\tau$ is a temperature parameter that controls the sharpness of the probability distribution, scaling the cosine similarities to enhance discriminability.

\subsubsection{Anomaly Mixture-of-Experts (AnomalyMoE)}
We propose a dense dual-gating mechanism, to effectively aggregate multi-scale experts. In contrast to sparse top-$k$ selection, our method utilizes differentiable soft-gating to adaptively combine all feature scales, simultaneously addressing inter-layer scale selection and inter-layer noise filtering. 
The detailed inference procedure is summarized in Algorithm \ref{alg:anomaly_moe}.

We posit that the global semantic context carries the most informative feature scale for the input image. To model this dependency, we employ the global synergistic feature $V_{syn}^{global}$ to generate scale-specific weights $w_{scale} \in \mathbb{R}^L$:
\begin{equation}
    w_{scale} = \text{Softmax}\left(\mathcal{F}_{gate}^{global}(V_{syn}^{global})\right)
    \label{eq:moe_scale} ,%
\end{equation}
where $\mathcal{F}_{gate}^{global}$ is a Multi-Layer Perceptron (MLP). This global gating branch allows the model to adaptively highlight specific layers, such as focusing on shallow layers when processing texture-centric images.
Complementing the scale selection, we recognize that even within the optimal layer, background regions may introduce irrelevant noise. To further purify the features, we employ a localized gating network $\mathcal{F}_{gate}^{local}$ (implemented as a $1\times1$ convolution) to generate a spatial attention mask $M_{spatial}^{(l)} \in [0,1]^{H \times W}$:
\begin{equation}
    M_{spatial}^{(l)} = \sigma\left(\mathcal{F}_{gate}^{local}(V_{syn}^{local, (l)})\right)
    \label{eq:moe_spatial} ,%
\end{equation}
where $\sigma$ denotes the sigmoid function. This local gating branch acts as a pixel-wise filter to suppress non-anomalous regions.
Finally, the fused pixel-wise anomaly map $\mathcal{P}_{map}$ is obtained by aggregating the experts, weighted by both the global scale importance and local spatial saliency:
\begin{equation}
    \mathcal{P}_{map} = \sum_{l=1}^{L} w_{scale}[l] \cdot \left( M_{spatial}^{(l)} \odot \mathcal{P}^{(l)}_{anom} \right) ,%
    \label{eq:moe_agg}
\end{equation}
where $\odot$ denotes element-wise multiplication. Through this dual-gated aggregation, the model effectively filters out irrelevant scales and background noise, yielding a high-quality anomaly map.

\begin{algorithm}[h]
\setstretch{0.9}
\caption{Anomaly Mixture-of-Experts (AnomalyMoE)}
\label{alg:anomaly_moe}
\SetAlgoLined
\KwIn{Multi-scale Local Features $\{V_{syn}^{local, (l)}\}_{l=1}^L$, Global Feature $V_{syn}^{global}$, Text Prompt $T_{final}$}
\KwParam{MoE Parameters $\Theta_{MoE} = \{\mathcal{F}_{gate}^{global}, \mathcal{F}_{gate}^{local}\}$}\;
\KwOut{Fused Anomaly Map $\mathcal{P}_{map}$}

\tcp{1. Generate Raw Experts (Layer-wise Maps)}
Initialize map list $\mathcal{P}_{experts} \leftarrow []$\;
\For{$l \leftarrow 1$ \KwTo $L$}{
    $\mathcal{M}^{(l)} \leftarrow \text{CosSim}(V_{syn}^{local, (l)}, T_{final})$ \tcp*[r]{Eq.~\ref{eq:raw_map}}
    $\mathcal{P}^{(l)} \leftarrow \text{Softmax}(\mathcal{M}^{(l)})$ \tcp*[r]{Eq.~\ref{eq:prob_map}}
    Add $\mathcal{P}^{(l)}$ to $\mathcal{P}_{experts}$\;
}

\tcp{2. Branch A: Global Scale Gating (Inter-layer Weighting)}
Scale Logits $h_{scale} \leftarrow \mathcal{F}_{gate}^{global}(V_{syn}^{global})$\;
Scale Weights $w_{scale} \leftarrow \text{Softmax}(h_{scale})$ \tcp*[r]{Eq.~\ref{eq:moe_scale}}

\tcp{3. Branch B: Local Spatial Gating (Intra-layer Attention)}
Initialize fused map $\mathcal{P}_{map} \leftarrow \mathbf{0}$\;
\For{$l \leftarrow 1$ \KwTo $L$}{
    Spatial Logits $H_{spatial}^{(l)} \leftarrow \mathcal{F}_{gate}^{local}(V_{syn}^{local, (l)})$\;
    Spatial Mask $M_{spatial}^{(l)} \leftarrow \sigma(H_{spatial}^{(l)})$ \tcp*[r]{Eq.~\ref{eq:moe_spatial}}
    
    \tcp{4. Dual-Gated Aggregation}
    $\mathcal{P}_{weighted}^{(l)} \leftarrow w_{scale}[l] \cdot (M_{spatial}^{(l)} \odot \mathcal{P}^{(l)})$\;
    $\mathcal{P}_{map} \leftarrow \mathcal{P}_{map} + \mathcal{P}_{weighted}^{(l)}$ \tcp*[r]{Eq.~\ref{eq:moe_agg}}
}

\Return $\mathcal{P}_{map}$
\end{algorithm}

\subsubsection{Score Enhancement Strategy}
We utilize the generated anomaly map to enhance the global classification score by extracting the most prominent local evidence via Spatial Max-Pooling:
\begin{equation}
    S_{local} = \max_{h,w} \mathcal{P}_{map}(h,w)  .%
\end{equation}
The local evidence acts as a dynamic gain term. We inject it into the global semantic score to refine the prediction. The result is the final anomaly score $S_{final}$:
\begin{equation}
    S_{final} = (1-\gamma) \cdot \langle V_{syn}^{global}, T_{final} \rangle + \gamma \cdot S_{local} ,%
    \label{eq:final_score}
\end{equation}
where $\gamma$ balances the contribution of global semantics and local details.

\subsection{Objective Function}
\label{subsec:optimization}

The SSVP framework is trained end-to-end using a joint objective that supervises both pixel-level segmentation and image-level classification. For pixel-level supervision, we employ the focal loss on the anomaly probability map $\mathcal{P}^{(l)}_{\text{anom}}$ from the VTAM module to address the severe imbalance between normal and anomalous pixels. Given the ground-truth mask $Y \in \{0,1\}^{H \times W}$, the segmentation loss is defined as:
\begin{equation}
\label{eq:focal_loss}
\begin{split}
    \mathcal{L}_{Seg} = - \frac{1}{HW} \sum_{h,w} \bigg( & Y_{h,w} (1-p_{h,w})^\lambda \log(p_{h,w}) \\
    & + (1-Y_{h,w}) p_{h,w}^\lambda \log(1-p_{h,w}) \bigg),
\end{split}
\end{equation}
where $p_{h,w} = \mathcal{P}^{(l)}_{anom}(h,w)$ denotes the predicted probability, and $\lambda$ is the focusing parameter.In addition to segmentation, image-level supervision is crucial to optimize the VTAM fusion mechanism and enforce the global representation to be anomaly-aware. We supervise the final fused score $S_{final}$ using the image-level binary label $y_{img} \in \{0, 1\}$ via the binary cross-entropy (BCE) loss:
\begin{equation}
    \mathcal{L}_{Class} = \text{BCE}(\sigma(S_{final}), y_{img}) ,%
    \label{eq:class_loss} 
\end{equation}
where $\sigma(\cdot)$ is the Sigmoid function used to normalize the score. By back-propagating $\mathcal{L}_{Class}$, the model learns to dynamically adjust the contribution of local and global features for robust decision-making.Consequently, the total training objective $\mathcal{L}_{total}$ aggregates the segmentation and classification losses with the regularization terms derived from the VCPG module:
\begin{equation}
    \mathcal{L}_{total} = \mathcal{L}_{Seg} + \mathcal{L}_{Class} + \lambda_1 \mathcal{L}_{VAE} + \lambda_2 \mathcal{L}_{reg} ,%
    \label{eq:total_loss}
\end{equation}
where $\mathcal{L}_{VAE}$ and $\mathcal{L}_{reg}$ enforce latent distribution and semantic consistency constraints, respectively, which are balanced by hyperparameters $\lambda_1$ and $\lambda_2$.The complete optimization process, including the calculation of these losses and the gradient update steps, is summarized in Algorithm \ref{alg:ssvp_opt}.

\begin{algorithm}[h]
\setstretch{0.9}
\caption{Optimization Strategy of SSVP}
\label{alg:ssvp_opt}
\SetAlgoLined
\SetKwInput{KwHyper}{Hyperparameters}
\KwIn{Training Dataset $\mathcal{D}$, Max Epochs $E$, Learning rate $\eta$}
\KwHyper{KL weight $\beta$, Loss weights $\lambda_1, \lambda_2$, Margin $\xi$}
\KwParam{Learnable Parameters $\Theta$ (Initialized)}\;
\KwOut{Optimized Parameters $\Theta^*$}

\For{epoch $= 1$ to $E$}{
    \For{batch $(I, GT_{local}, GT_{global}) \in \mathcal{D}$}{
        \tcp{1. Execute Forward Pipeline (Algorithm \ref{alg:ssvp_forward})}
        $\mathcal{P}_{map}, S_{final}, \Phi_{aux} \leftarrow \text{RunForward}(I; \Theta)$\;
        Unpack $\Phi_{aux} \to \{z, V_{syn}^{global}, T_{init}, T_{final}\}$\;
        
        \tcp{2. Supervision Losses}
        $\mathcal{L}_{sup} \leftarrow \text{Focal}(\mathcal{P}_{map}, GT_{local}) + \text{BCE}(S_{final}, GT_{global})$ 
        \tcp*[r]{Eq.~\ref{eq:focal_loss}, \ref{eq:class_loss}}
        
        \tcp{3. Image Reconstruction Distribution Constraint}
        Reconstruct Feature: $\hat{v} \leftarrow D_\theta(z)$\;
        $\mathcal{L}_{VAE} \leftarrow \|V_{syn}^{global} - \hat{v}\|_2^2 + \beta D_{KL}(q_\phi(z|V_{syn}^{global}) \| \mathcal{N}(0, \mathbf{I}))$ \tcp*[r]{Eq.~\ref{eq:vae_loss}}
        
        \tcp{4. Text Semantic Consistency}
        $\mathcal{L}_{reg} \leftarrow \text{MarginLoss}(T_{final}, T_{init}; \xi)$ \tcp*[r]{Eq.~\ref{eq:reg_loss}}
        
        \tcp{5. Total Objective and Gradient Update}
        $\mathcal{L}_{total} \leftarrow \mathcal{L}_{sup} + \lambda_1\mathcal{L}_{VAE} + \lambda_2\mathcal{L}_{reg}$ \tcp*[r]{Eq.~\ref{eq:total_loss}}
        $\Theta \leftarrow \Theta - \eta \nabla_{\Theta} \mathcal{L}_{total}$ \tcp*[r]{Gradient Descent}
    }
}
\Return $\Theta$
\end{algorithm}

\section{Experiments}
\label{sec:experiments}

In this section, we conduct extensive experiments to evaluate the effectiveness of the proposed SSVP framework. We first introduce the datasets and evaluation metrics. Subsequently, we describe the implementation details and training protocols. We then compare SSVP with state-of-the-art (SOTA) zero-shot anomaly detection methods, followed by in-depth ablation studies and qualitative analysis to verify the contribution of each component.

\subsection{Datasets and Evaluation Metrics}
\label{subsec:datasets}

To comprehensively evaluate the robustness and generalization capability of SSVP across diverse industrial scenarios, we utilize seven challenging benchmarks covering object-centric, texture-centric, and real-world surface defect scenarios:
\begin{itemize}
    \item MVTec-AD \citep{ref_mvtec}: A standard industrial benchmark comprising 5,354 high-resolution images across 15 categories (5 textures, 10 objects). It features varied defect types ranging from subtle scratches to structural deformations.
    \item VisA \citep{ref_visa}: A large-scale dataset with 9,621 images spanning 12 categories. It includes objects with complex structures, multiple instances, and loose spatial layouts, posing significant challenges for zero-shot localization.
    \item BTAD \citep{ref_btad}: The BeanTech Anomaly Detection dataset contains 2,540 images of 3 real-world industrial products, focusing on subtle surface defects.
    \item KSDD2 \citep{ref_ksdd2}: The Kolektor Surface-Defect Dataset 2 consists of 3,355 images of electrical commutators, specifically designed for detecting fine surface cracks.
    \item RSDD \citep{ref_rsdd}: The Rail Surface Defect Dataset contains Type-I and Type-II rail surface images, challenging models with complex background noise.
    \item DAGM \citep{ref_dagm}: A synthetic dataset (DAGM 2007) containing 10 classes of varying textures with generated anomalies, used to evaluate robustness on texture defects.
    \item DTD-Synthetic \citep{ref_dtd}: Based on the Describable Textures Dataset, we evaluate on synthetic anomalies to test the model's ability to handle diverse texture patterns.
\end{itemize}


We adopt a strict Cross-Domain Zero-Shot Transfer protocol. Specifically, we use the training set of MVTec-AD as the auxiliary source domain for VisA evaluation, and the training set of VisA as the source domain for all other datasets (MVTec-AD, BTAD, etc.), ensuring no target data leakage.

\textbf{Comparison Baselines.} We benchmark SSVP against state-of-the-art methods in the ZSAD domain to verify its superiority. The comparison models include: (1) Zero-Shot Baselines: WinCLIP \citep{ref_winclip}, APRIL-GAN \citep{ref_aprilgan}; (2) Dynamic Prompting Methods: AnomalyCLIP \citep{ref_anomalyclip}, AdaCLIP \citep{ref_adaclip} and  Bayes-PFL \citep{ref_bayes_pfl}.

\textbf{Zero-Shot Training Protocol.} We adopt a strict Cross-Domain Zero-Shot Transfer protocol to ensure no data leakage from the target domain.
\begin{itemize}
    \item For VisA Evaluation: We utilize the training sets of the 15 categories from MVTec-AD as the auxiliary source domain to optimize the learnable parameters (Prompt Banks, VAE, HSVS weights).
    \item For Other Datasets: For evaluations on MVTec-AD, BTAD, KSDD2, RSDD, DAGM, and DTD-Synthetic, we utilize the training sets of the 12 categories from VisA as the auxiliary source domain.
\end{itemize}
Under this setting, the model strictly adheres to the zero-shot assumption, as it never encounters any normal or abnormal samples (images or text descriptions) from the target categories during the training phase.

\textbf{Evaluation Metrics.} We employ standard metrics to evaluate performance at both image and pixel levels:
\begin{itemize}
    \item Image-level Metrics: We report AUROC, F1-Max (maximum F1-score at optimal threshold), and AP (Average Precision) to comprehensively assess anomaly classification performance, ensuring a robust evaluation of the model's ability to distinguish between normal and anomalous samples across various decision thresholds.
    \item Pixel-level Metrics: To evaluate the fine-grained defect localization quality, we report AUROC, PRO (Per-Region Overlap), and AP. Notably, PRO scales with the overlap ratio of connected components, making it more sensitive to the precise coverage of defect regions compared to pixel-wise AUROC, thereby treating anomalies of varying sizes with equal importance and avoiding bias from large background areas.
\end{itemize}

\subsection{Implementation Details}
\label{subsec:implementation}

Architecture and Feature Alignment. We utilize CLIP (ViT-L/14) and DINOv3 (ViT-L/16) as the frozen backbone encoders. To achieve pixel-level alignment across modalities, we adopt a Resolution Adaptation Strategy to counteract the discrepancy in patch sizes. Specifically, given that CLIP uses a patch size of 14, we resize the input image to $518 \times 518$ for the CLIP branch. Conversely, as DINOv3 operates with a patch size of 16, we set the input resolution to $592 \times 592$ for the DINO branch. This configuration ensures that the feature maps from both encoders share an identical spatial grid size of $H' \times W' = 37 \times 37$, facilitating direct interaction without interpolation artifacts.

To capture hierarchical semantic and structural information, we extract multi-scale features for fusion. We select indices $\{6, 12, 18, 24\}$ from CLIP and $\{3, 6, 9, 11\}$ from DINOv3 as inputs to the HSVS module. The dimension of the semantic manifold is projected to $D_c=768$. For the VCPG module, we employ 6 learnable prompts, comprising 3 normal prompts and 3 abnormal prompts.

To ensure fairness, all models employ the same training configuration parameters, with the epoch set to 15 and the batch size set to 8. All experiments are conducted on a single NVIDIA RTX 4090 GPU (24GB) using PyTorch. The initial learning rates are set to $5 \times 10^{-4}$ for the prompt embeddings and $1 \times 10^{-4}$ for the VAE and projection layers, following a cosine decay scheduler. Regarding loss hyper-parameters, we set $\beta=0.1$ for the KL divergence term, $\lambda_1=1.0$ for the VAE reconstruction, and $\lambda_2=0.5$ for the semantic regularization. The threshold $\xi$ in Eq. \ref{eq:reg_loss} is set to 0.85, and the VTAM balancing factor $\gamma$ is set to 0.5.


\subsection{Comparative Experiments}
\label{subsec:comparison}

We compare SSVP with leading ZSAD methods, including: 
(1) Static Prompting Methods: WinCLIP \citep{ref_winclip}, APRIL-GAN \citep{ref_aprilgan}; 
(2) Dynamic Prompting Methods: AnomalyCLIP \citep{ref_anomalyclip}, AdaCLIP \citep{ref_adaclip}, Bayes-PFL \citep{ref_bayes_pfl}.

Quantitative Results. As shown in Table \ref{tab:comparison_industrial}, SSVP sets a new SOTA across seven benchmarks. Notably, it surpasses Bayes-PFL on MVTec-AD with 93.0\% Image-AUROC and 92.2\% Pixel-AUROC. SSVP demonstrates robustness by dominating complex scenarios (VisA, RSDD, DAGM) and achieving the best pixel-level metrics on five datasets. Crucially, the consistent superiority in the PRO metric validates that our HSVS module effectively captures fine-grained structural anomalies often missed by other CLIP methods.

\begin{table}[h]
\centering
\caption{Comparison with existing state-of-the-art methods in the Industrial domain. The best results are highlighted in \best{bold}, while the second-best are \second{underlined}.}
\label{tab:comparison_industrial}
\renewcommand{\arraystretch}{1.4}  
\setlength{\tabcolsep}{3pt}      
\small
\resizebox{\textwidth}{!}{
\begin{tabular}{cccccccc}
\toprule
\textbf{Metric} & \textbf{Dataset} & WinCLIP  & APRIL-GAN  & AnomalyCLIP & AdaCLIP  & Bayes-PFL  & \textbf{SSVP (Ours)} \\
\midrule
\multirow{7}{*}{\shortstack{Image-level\\(AUROC, F1-Max,\\AP)}} 
  & MVTec-AD & ({88.7}, {92.6}, {94.8}) & ({85.7}, {90.2}, {93.3}) & ({89.3}, {92.4}, {96.0}) & ({90.7}, {92.6}, \second{96.2}) & (\second{92.0}, \second{93.0}, {96.0}) & (\best{93.0}, \best{93.3}, \best{96.7}) \\
  & VisA     & ({77.6}, {78.8}, {77.1}) & ({77.9}, {78.4}, {81.1}) & ({82.0}, {80.1}, {85.2}) & ({82.6}, {81.3}, {84.4}) & (\second{87.1}, \second{83.4}, \second{89.0}) & (\best{88.2}, \best{85.4}, \best{90.4}) \\
  & BTAD     & ({83.3}, {81.0}, {84.1}) & ({74.0}, {69.5}, {71.2}) & ({89.0}, {85.5}, {90.8}) & (\second{91.6}, \best{90.1}, {92.6}) & ({91.4}, {88.6}, \best{94.2}) & (\best{94.2}, \second{89.8}, \second{93.6}) \\
  & KSDD2    & ({93.2}, {71.1}, {77.4}) & ({90.0}, {69.7}, {74.1}) & ({91.6}, {70.8}, {77.4}) & ({95.4}, {84.2}, {95.6}) & (\best{97.0}, \best{92.1}, \best{97.6}) & (\second{96.9}, \second{91.8}, \second{97.5}) \\
  & RSDD     & ({85.3}, {73.5}, {65.3}) & ({73.1}, {59.7}, {50.5}) & ({73.5}, {59.0}, {55.0}) & (\second{92.1}, {75.0}, \second{70.8}) & ({90.3}, \second{85.6}, {70.1}) & (\best{98.5}, \best{95.2}, \best{98.6}) \\
  & DAGM     & ({89.6}, {86.4}, {90.4}) & ({90.4}, {86.9}, {90.1}) & ({95.6}, {93.2}, {94.6}) & ({96.5}, {94.1}, {95.7}) & (\second{97.7}, \second{95.7}, \second{97.0}) & (\best{98.0}, \best{96.2}, \best{97.4}) \\
  & DTD-Syn  & ({93.9}, {94.2}, \best{98.1}) & ({83.7}, {89.3}, {93.5}) & (\best{94.5}, {94.5}, {97.7}) & ({92.8}, {92.2}, {97.0}) & ({93.8}, \best{95.1}, {97.8}) & (\second{94.0}, \second{94.6}, \second{98.0}) \\
\cmidrule{1-8}
\multirow{7}{*}{\shortstack{Pixel-level\\(AUROC, PRO,\\AP)}} 
  & MVTec-AD & ({82.5}, {64.2}, {18.1}) & ({88.9}, {43.6}, {44.7}) & ({88.9}, {81.4}, {34.5}) & ({85.5}, {40.8}, {43.1}) & (\second{91.9}, \second{87.8}, \second{48.3}) & (\best{92.2}, \best{89.0}, \best{50.0}) \\
  & VisA     & ({79.3}, {56.2}, {4.9})  & ({94.1}, {86.4}, {25.2}) & (\second{95.2}, {86.4}, {21.0}) & ({94.7}, {70.8}, {26.4}) & ({93.2}, \second{89.2}, \second{29.7}) & (\best{96.2}, \best{90.5}, \best{30.1}) \\
  & BTAD     & ({68.4}, {31.8}, {10.1}) & (\second{91.3}, {43.0}, {32.9}) & ({90.2}, {43.3}, {38.2}) & ({87.1}, {15.8}, {34.3}) & ({90.2}, \second{73.3}, \second{45.1}) & (\best{96.5}, \best{84.7}, \best{51.2}) \\
  & KSDD2    & ({92.4}, {90.2}, {14.3}) & ({93.4}, {50.1}, {59.7}) & ({98.7}, {91.5}, {40.2}) & ({95.1}, {68.2}, {40.4}) & (\best{99.6}, \second{97.1}, \best{72.3}) & (\second{99.5}, \best{98.8}, \second{71.5}) \\
  & RSDD     & ({93.8}, {74.4}, {2.2})  & (\second{99.4}, {65.2}, {27.6}) & ({97.8}, \second{93.2}, {18.7}) & ({93.6}, {48.3}, {37.1}) & (\second{99.4}, {92.1}, \second{39.1}) & (\best{99.7}, \best{98.7}, \best{43.4}) \\
  & DAGM     & ({82.5}, {55.0}, {3.0})  & ({98.6}, {42.9}, {41.6}) & ({98.1}, \second{95.6}, {28.5}) & ({95.4}, {39.1}, \second{45.2}) & (\second{98.7}, {94.6}, {42.5}) & (\best{99.4}, \best{98.4}, \best{50.5}) \\
  & DTD-Syn  & ({82.3}, {55.2}, {11.5}) & ({96.4}, {41.5}, {63.1}) & ({97.1}, {88.2}, {52.1}) & ({94.0}, {24.7}, {52.4}) & (\second{97.2}, \second{94.1}, \best{69.9}) & (\best{98.1}, \best{94.3}, \second{64.6}) \\
\bottomrule
\end{tabular}
}
\end{table}

Class-wise Performance. Table \ref{tab:mvtec_breakdown} details results on MVTec-AD, showing that SSVP performs consistently across categories. For Carpet, Grid and Leather, it achieves 100\% Image-AUROC, indicating that DINOv3's high-frequency details are well preserved and aligned with text semantics. For complex structured objects, SSVP also improves notably on Cable and Metal Nut, it outperforms the strongest baseline by 2.0\% and 1.6\% in Pixel-AUROC, respectively. This demonstrates that VCPG effectively directs attention to anomaly-sensitive regions, suppressing background noise.

\begin{table}[H]
\centering
\caption{Class-wise comparison on MVTec-AD. We report Image-level ($\text{I}_{\text{L}}$) and Pixel-level ($\text{P}_{\text{L}}$) AUROC (\%) scores. The best results are highlighted in \best{bold}, and second-best are \second{underlined}.}
\label{tab:mvtec_breakdown}
\tiny
\renewcommand{\arraystretch}{1.0} 
\setlength{\tabcolsep}{1pt}      

\resizebox{1.0\textwidth}{!}{
\begin{tabular}{
    c c  
    *{10}{x{20pt}} 
}
\toprule
\multirow{2}{*}{Category} & \multirow{2}{*}{Class} & 
\multicolumn{2}{c}{WinCLIP } & 
\multicolumn{2}{c}{AnomalyCLIP } & 
\multicolumn{2}{c}{AdaCLIP } & 
\multicolumn{2}{c}{Bayes-PFL } & 
\multicolumn{2}{c}{\textbf{SSVP}} \\

\cmidrule(lr){3-4} \cmidrule(lr){5-6} \cmidrule(lr){7-8} \cmidrule(lr){9-10} \cmidrule(lr){11-12}

 & & $\text{I}_{\text{L}}$ & $\text{P}_{\text{L}}$ & $\text{I}_{\text{L}}$ & $\text{P}_{\text{L}}$ & $\text{I}_{\text{L}}$ & $\text{P}_{\text{L}}$ & $\text{I}_{\text{L}}$ & $\text{P}_{\text{L}}$ & $\text{I}_{\text{L}}$ & $\text{P}_{\text{L}}$ \\
\midrule

\multirow{5}{*}{Textures} 
 & Carpet     & {96.5} & {94.1} & {95.2} & \second{95.5} & \second{98.2} & {88.5} & \best{100} & \best{99.5} & \best{100} & \best{99.5} \\
 & Grid       & {89.8} & {92.3} & {93.1} & {93.4} & {95.5} & {86.2} & \second{99.6} & \second{98.5} & \best{100} & \best{99.1} \\
 & Leather    & {97.2} & {96.5} & {97.5} & {97.2} & \second{99.1} & {92.4} & \best{100} & \second{99.5} & \best{100} & \best{99.6} \\
 & Tile       & {94.5} & {85.4} & {93.0} & {91.0} & {96.8} & {89.5} & \second{99.2} & \second{94.2} & \best{99.9} & \best{94.9} \\
 & Wood       & {95.2} & {87.2} & {92.5} & {90.5} & {96.2} & {90.1} & \second{99.3} & \best{97.7} & \best{99.6} & \second{97.5} \\
\midrule

\multirow{10}{*}{Objects} 
 & Bottle     & {95.0} & {88.5} & {94.8} & {93.5} & {95.5} & {91.2} & \second{95.9} & \second{93.8} & \best{96.0} & \best{94.4} \\
 & Cable      & {81.5} & {73.2} & {82.4} & {75.2} & \second{83.6} & {75.5} & {81.6} & \second{77.1} & \best{85.6} & \best{77.4} \\
 & Capsule    & {89.2} & {81.5} & {90.5} & {95.8} & {91.0} & {88.4} & \second{91.5} & \second{96.1} & \best{96.0} & \best{96.2} \\
 & Hazelnut   & {92.4} & {84.0} & {94.2} & {95.5} & \second{95.8} & {92.6} & {93.0} & \second{98.6} & \best{98.5} & \best{98.8} \\
 & Metal Nut  & {76.8} & {68.5} & \second{77.2} & {73.5} & \second{77.2} & {71.8} & {76.1} & \second{74.4} & \best{78.8} & \best{74.8} \\
 & Pill       & {86.5} & {83.4} & {87.5} & {86.5} & \second{88.4} & {85.2} & {85.6} & \second{87.0} & \best{89.2} & \best{87.4} \\
 & Screw      & {80.2} & {74.1} & {81.8} & {90.8} & {83.5} & {82.5} & \best{90.4} & \best{98.8} & \second{87.0} & \second{98.6} \\
 & Toothbrush & {82.5} & {78.2} & {84.1} & {91.2} & {84.8} & {86.4} & \best{88.9} & \best{95.2} & \second{85.5} & \best{96.3} \\
 & Transistor & {77.8} & {63.5} & {78.5} & {68.4} & {78.0} & {69.1} & \best{80.2} & \second{69.7} & \second{79.1} & \best{70.4} \\
 & Zipper     & {95.9} & {87.1} & {96.7} & {94.9} & {95.5} & {92.6} & \best{99.3} & \second{98.3} & \best{99.3} & \best{98.4} \\
\midrule

\multicolumn{2}{l}{\textit{Average}} & {88.7} & {82.5} & {89.3} & {88.9} & {90.7} & {85.5} & \second{92.0} & \second{91.9} & \best{93.0} & \best{92.2} \\
\bottomrule
\end{tabular}
}
\end{table}


\subsection{Visualization Analysis}
\label{subsec:visualization}

This section visualizes the proposed fusion mechanism's effectiveness and generalization via SOTA comparisons, ablations, and diverse benchmarks.

\begin{figure}[H]
    \centering
    \includegraphics[width=1\linewidth, trim=15mm 90mm 60mm 90mm, clip]{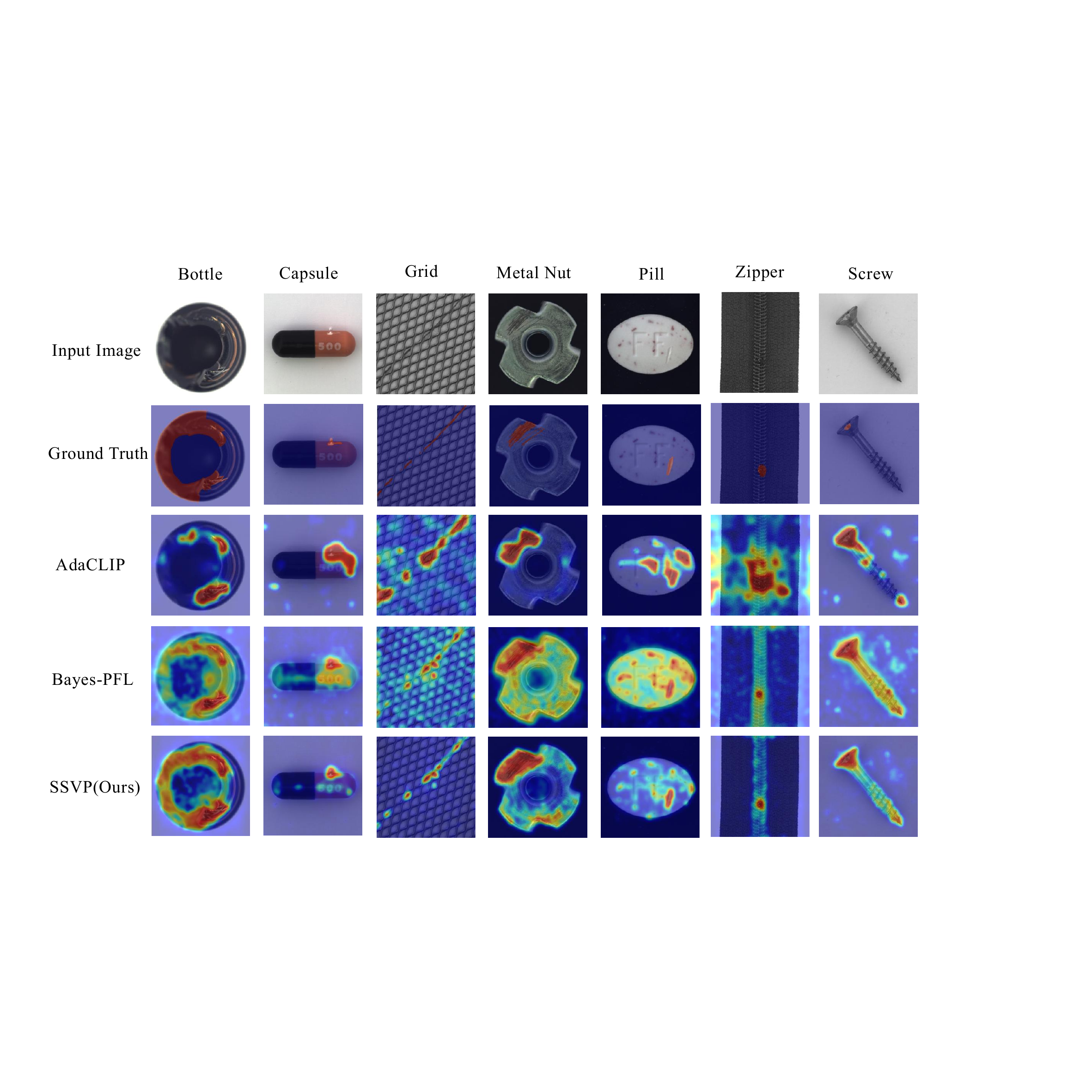}
    \caption{Qualitative comparison with SOTA methods. Visualization of anomaly heatmaps shows that SSVP produces significantly stronger responses on anomalous regions and achieves clearer foreground-background separation compared to standard ZSAD methods.}
    \label{fig:vis_qualitative}
\end{figure}

Qualitative Anomaly Localization. Figure \ref{fig:vis_qualitative} visualizes the segmentation results on typical industrial samples. Compared to AdaCLIP and Bayes-PFL, SSVP generates precise anomaly maps with sharp boundaries. This precision is primarily attributed to the HSVS module, which successfully preserves high-frequency structural details.

\begin{figure}[H]
    \centering
    \includegraphics[width=1\linewidth, trim=20mm 25mm 120mm 30mm, clip]{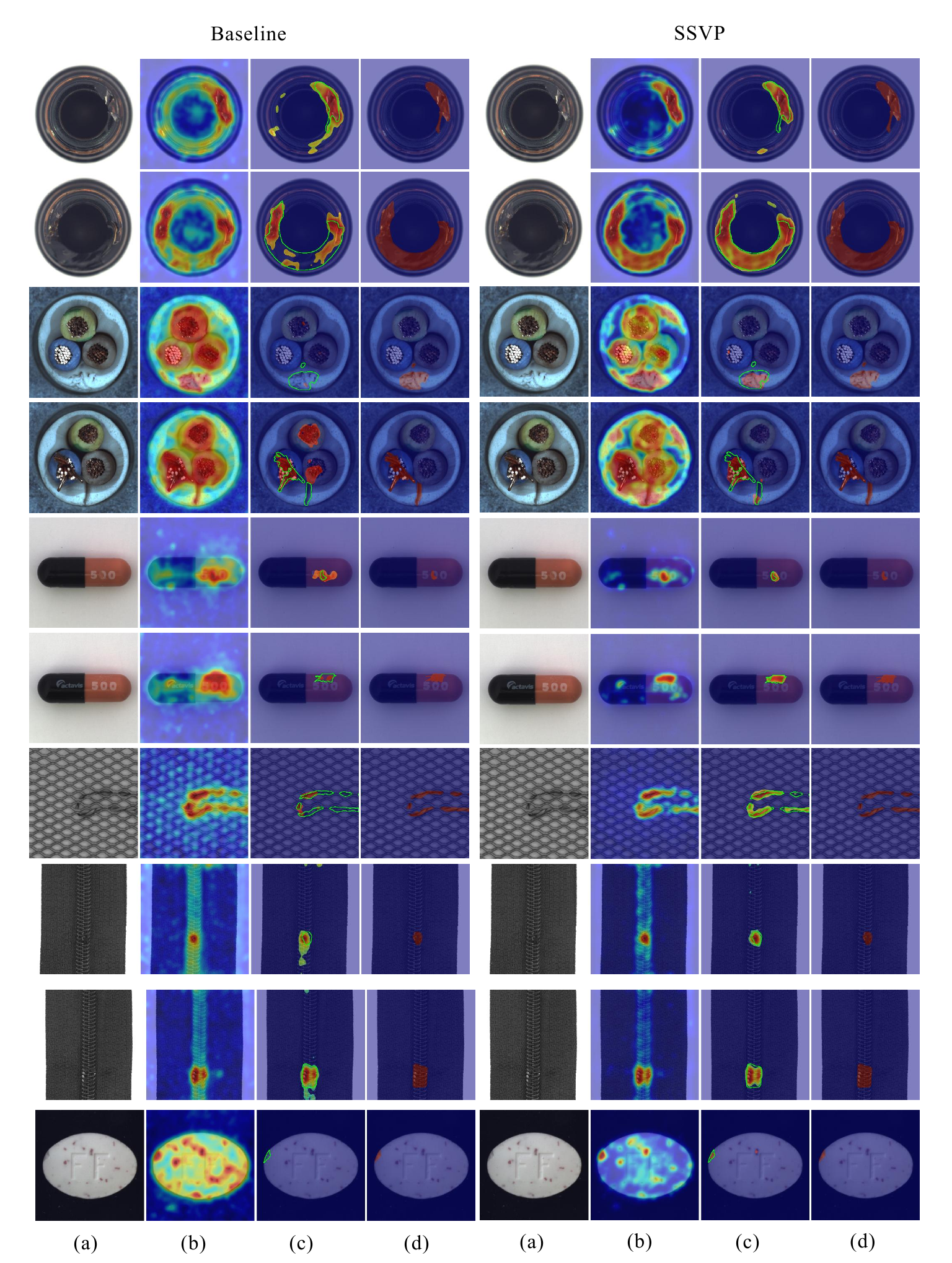}
    \caption{Visual comparison of anomaly localization between Baseline and SSVP. The columns display: (a) Original input images; (b) The generated anomaly score maps; (c) Feature maps processed by threshold filtering, where ground-truth areas are outlined by green contours; and (d) The ground-truth masks, indicated by red regions.}
    \label{fig:vis_baseline_comp}
\end{figure}

Effectiveness of Proposed Modules. To further verify the gain from our specific designs, we conduct a direct comparison between our full model and the baseline model in Figure \ref{fig:vis_baseline_comp}. As observed, after integrating our proposed components (HSVS and VCPG), the segmentation becomes significantly more accurate. Specifically, our method effectively suppresses false positives in complex background areas where the baseline often exhibits noise. 
SSVP demonstrates superior sensitivity to small-scale targets, localizing micro-defects that the baseline fails to capture. The resulting anomaly maps show a clearer separation between foreground defects and normal backgrounds, validating the effectiveness of our semantic-visual alignment strategy.

\begin{figure}[h]
    \centering
    \includegraphics[width=1\linewidth, trim=0mm 190mm 0mm 70mm, clip]{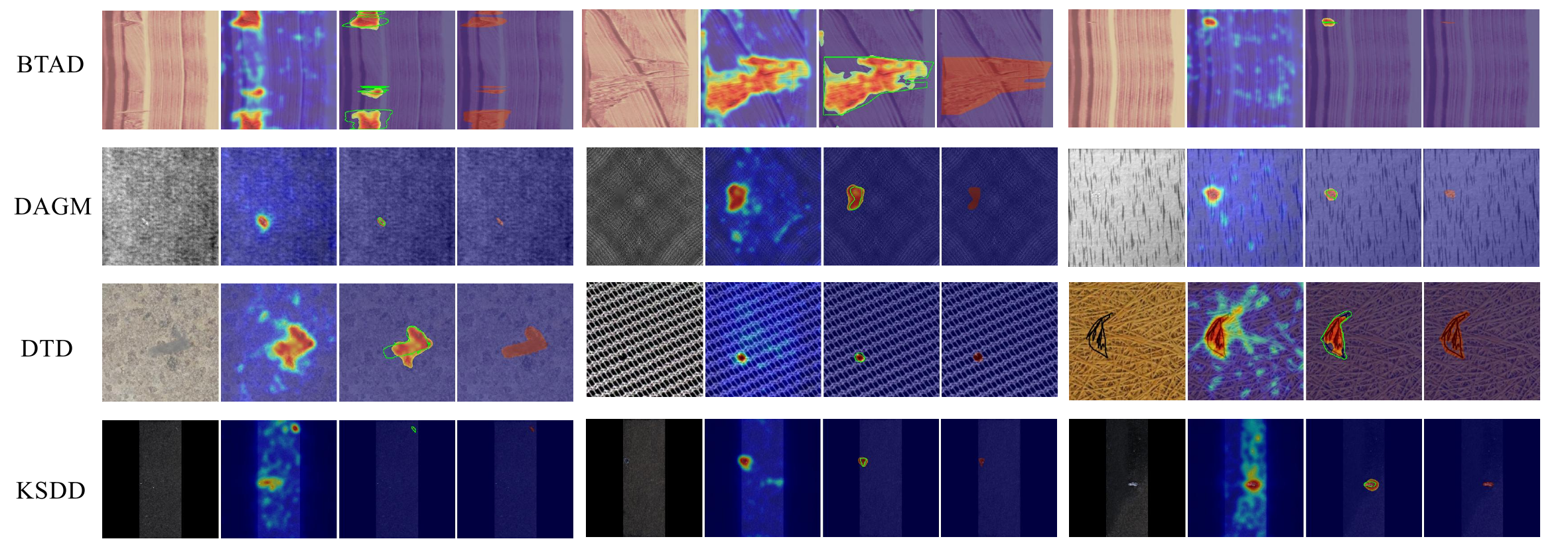}
    \caption{Visualization of ZSAD results across multiple datasets. We visualize sample results from BTAD, DAGM, DTD-Synthetic, and KSDD2. The column layout is identical to Figure \ref{fig:vis_baseline_comp}: Original images; Anomaly score maps; Thresholded features with GT contours; and GT red masks.}
    \label{fig:vis_datasets}
\end{figure}

We further evaluate the robustness of SSVP on four additional challenging benchmarks: BTAD, DAGM, DTD-Synthetic, and KSDD2, as shown in Figure \ref{fig:vis_datasets}. The results verify the strong generalization capability of our framework. Specifically, SSVP adapts well to texture-centric scenarios, accurately identifying anomalies amidst complex patterns. For object-centric datasets with minute defects, the model maintains high sensitivity to small targets. This consistent performance across varying domains confirms that SSVP effectively learns transferable visual-semantic features.

\subsection{Ablation Studies}
\label{subsec:ablation}

We conduct component-wise ablation studies on the MVTec-AD dataset to systematically validate the contribution of each module. 
\begin{table}[h]
    \centering
    \caption{Component-wise ablation study on MVTec-AD. Baseline denotes Bayes-PFL \citep{ref_bayes_pfl}. Best results are in \best{bold}.}
    \label{tab:ablation}
    \scalebox{0.85}{ 
    \begin{tabular}{c|ccc|cc}
        \toprule
        \textbf{Model} & \textbf{HSVS} & \textbf{VCPG} & \textbf{VTAM} & \textbf{I-AUC} & \textbf{P-AUC} \\
        \midrule
        A (Baseline) & \ding{55} & \ding{55} & \ding{55} & 91.3 $\pm$ 0.6 & 91.2 $\pm$ 0.4 \\
        B & \ding{51} (Concat) & \ding{55} & \ding{55} & 90.8 $\pm$ 0.3 & 91.0 $\pm$ 0.4 \\
        C & \ding{51} (ATF) & \ding{55} & \ding{55} & 91.4 $\pm$ 0.2 & 92.1 $\pm$ 0.3 \\
        D & \ding{51} & \ding{51} & \ding{55} & {92.6 $\pm$ 0.3} & {92.5 $\pm$ 0.3} \\
        \textbf{E (Full)} & \ding{51} & \ding{51} & \ding{51} & \best{93.0 $\pm$ 0.2} & \best{92.6 $\pm$ 0.4} \\
        \bottomrule
    \end{tabular}
    }
\end{table}

We evaluate components in Table \ref{tab:ablation}. Model B's naive concatenation drops Image-AUROC to 90.8\% due to misalignment. HSVS in Model C aligns features, boosting Pixel-AUROC to 92.1\%. Adding VCPG in Model D reaches 92.7\% Pixel-AUROC, enhancing fine-grained sensitivity. Full SSVP (Model E) with VTAM calibrates scores for 93.0\% Image-AUROC and robust performance.

To quantitatively assess the quality of semantic-visual alignment, we introduce Text-Image Response (TIR) and Semantic Consistency Distance (SCD). TIR-FG measures the activation intensity on defect regions:
\begin{equation}
    \text{TIR}_{\text{FG}} = \frac{1}{|\Omega_{fg}|} \sum_{u \in \Omega_{fg}} \text{CosSim}(\mathbf{v}_u, \mathbf{t}_{abn}) ,%
\end{equation}
where $\Omega_{fg}$ denotes the ground-truth anomaly mask. SCD evaluates the fidelity of learned prompts relative to static prototypes:
\begin{equation}
    \text{SCD} = \| \mathbf{t}_{final} - \mathbf{t}_{static} \|_2 .%
\end{equation}

As presented in Table \ref{tab:semantic_analysis}, SSVP increases TIR-FG by a relative margin of 28.88\% compared to the baseline, validating that our dynamic prompts effectively anchor to discriminative visual defects. The SCD metric decreases for both normal and abnormal prompts, indicating that our margin-based regularization effectively prevents semantic drift and maintains category fidelity.

\begin{table}[H]
    \centering
    \caption{{Analysis of semantic-visual alignment.} Higher TIR-FG and lower SCD indicate better performance.}
    \label{tab:semantic_analysis}
    \scalebox{0.85}{
    \begin{tabular}{l ccc c}
        \toprule
        \multirow{2}{*}{\textbf{Metric}} & \textbf{Baseline} & \textbf{SSVP} & \multicolumn{2}{c}{\textbf{Improvement}} \\
        \cmidrule(lr){4-5}
        & (Identity) & (\textbf{Full}) & $\Delta$ Value & Relative Gain \\
        \midrule
        \textbf{TIR-FG} ($\uparrow$) & 0.3434 & \textbf{0.4425} & +0.0992 & {\textbf{28.88\%}} \\
        \midrule
        \textbf{SCD-Normal} ($\downarrow$) & 1.2863 & \textbf{1.2764} & -0.0099 & {\textbf{0.77\%}} \\
        \textbf{SCD-Abnormal} ($\downarrow$) & 1.3087 & \textbf{1.2804} & -0.0282 & {\textbf{2.16\%}} \\
        \bottomrule
    \end{tabular}
    }
\end{table}

Table~\ref{tab:ablation_backbone_scale} evaluates visual backbones. DINOv3 (self-supervised) outperforms reconstruction-based MAE (93.0\% vs. 90.5\% Image-AUROC). Scaling DINOv3 from ViT-Small to ViT-Large yields consistent gains, confirming that larger models provide richer structural details for anomaly detection.

\begin{table}[h]
    \centering
    \caption{{Impact of visual backbones.} Comparisons of pre-training paradigms and model scaling.}
    \label{tab:ablation_backbone_scale}
    \scalebox{0.85}{
    \begin{tabular}{l c c c c}
        \toprule
        \textbf{Auxiliary Model} & \textbf{Backbone} & \textbf{Params (M)} & \textbf{I-AUC (\%)} & \textbf{P-AUC (\%)} \\
        \midrule
        MAE \citep{ref_mae} & ViT-Large & 303 & 90.5 & 90.1 \\ 
        DINOv2 \citep{ref_dinov2} & ViT-Large & 304 & 91.1 & 91.0 \\ 
        \textbf{DINOv3 \citep{ref_dinov3}} & \textbf{ViT-Large} & \textbf{303} & \textbf{93.0} & \textbf{92.2} \\
        \midrule
        DINOv3 & ViT-Small & 29 & 90.9 & 90.8 \\ 
        DINOv3 & ViT-Base  & 86 & 92.2 & 92.0 \\ 
        \textbf{DINOv3} & \textbf{ViT-Large} & \textbf{303} & \textbf{93.0} & \textbf{92.2} \\
        \bottomrule
    \end{tabular}
    }
\end{table}

\begin{figure}[H]
    \centering
    \includegraphics[width=0.5\textwidth, trim=0mm 60mm 0mm 0mm, clip]{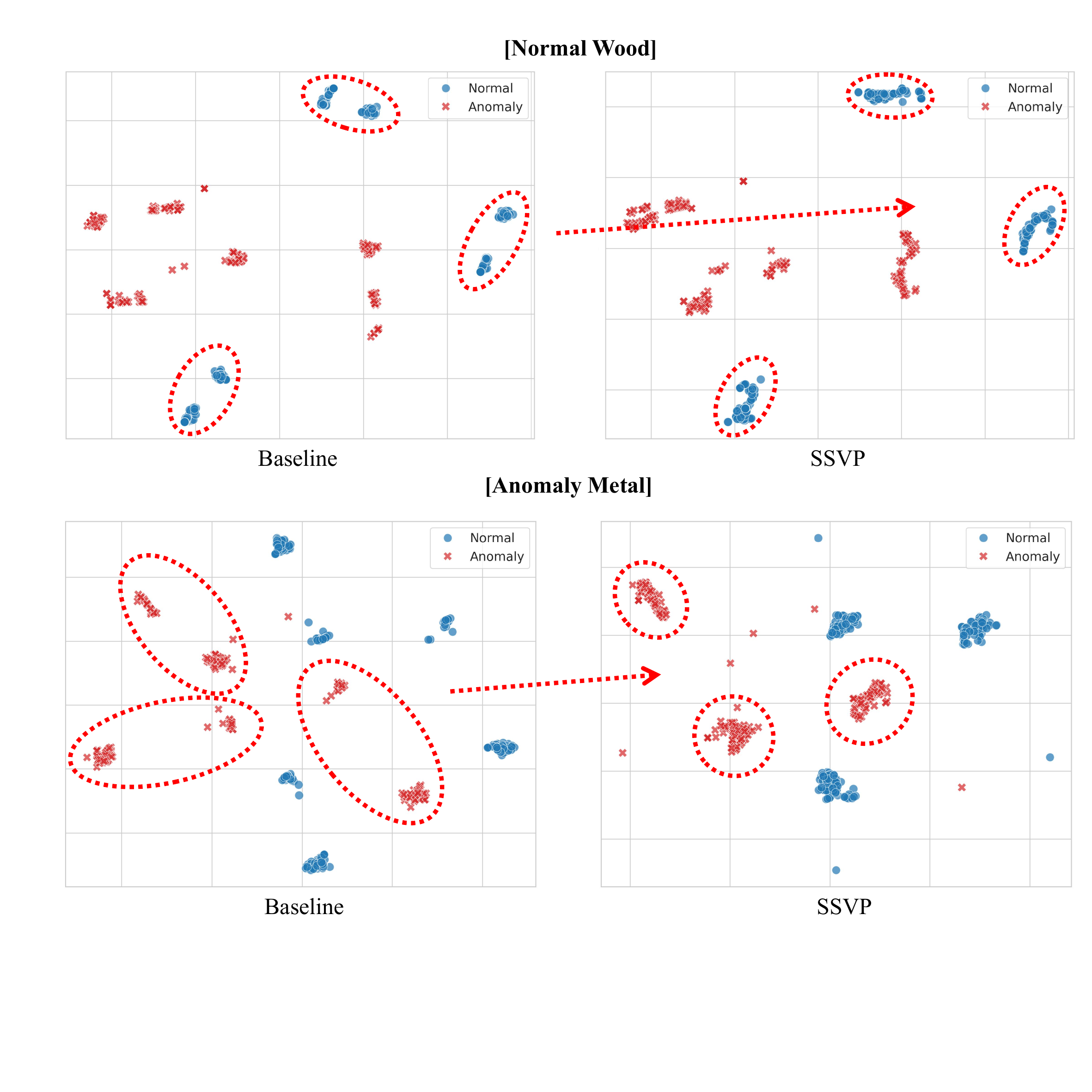}
    \caption{The t-SNE visualization of prompt embeddings on the BTAD dataset. The results reveal that SSVP generates a continuous and coherent semantic manifold compared to the fragmented baseline, effectively balancing semantic consistency with diversity.}
    \label{fig:vis_tsne}
\end{figure}

Figure~\ref{fig:vis_tsne} presents t-SNE embeddings on the BTAD dataset. Compared to the disconnected clusters in the baseline distribution, SSVP forms a coherent manifold, confirming that our vision-conditioned mechanism maintains semantic consistency. The distinct arrangement of prompt groups further ensures both diversity for capturing varied anomalies and alignment with class semantics.

\subsection{Parameter Analysis and Discussion}
\label{subsec:parameter}

To rigorously determine the optimal hyperparameters for SSVP, we employed a step-wise optimization strategy. 
We fix $\mathcal{P}^*\!=\!\{\gamma=0.5,\,\xi=0.85\}$ via coordinate-wise search and vary one parameter at a time: $\gamma$ (Eq.\,\eqref{eq:final_score}) balances global semantics against local anomaly evidence; $\xi$ (Eq.\,\eqref{eq:reg_loss}) sets the cosine-similarity floor between $T_{final}$ and $T_{init}$ to control semantic drift.

\begin{figure}[H]
    \centering
    \begin{minipage}{0.48\linewidth}
        \centering
        \includegraphics[width=\linewidth]{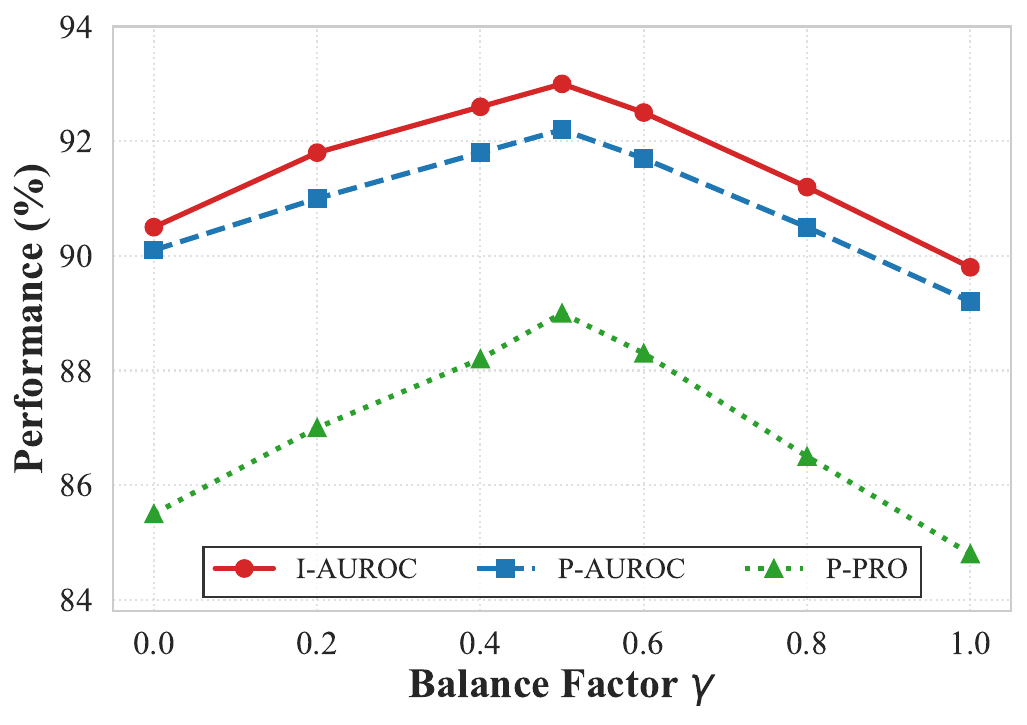}
        \caption{Analysis of the balancing factor $\gamma$.}
        \label{fig:gamma_analysis}
    \end{minipage}
    \hfill 
    \begin{minipage}{0.48\linewidth}
        \centering
        \includegraphics[width=\linewidth]{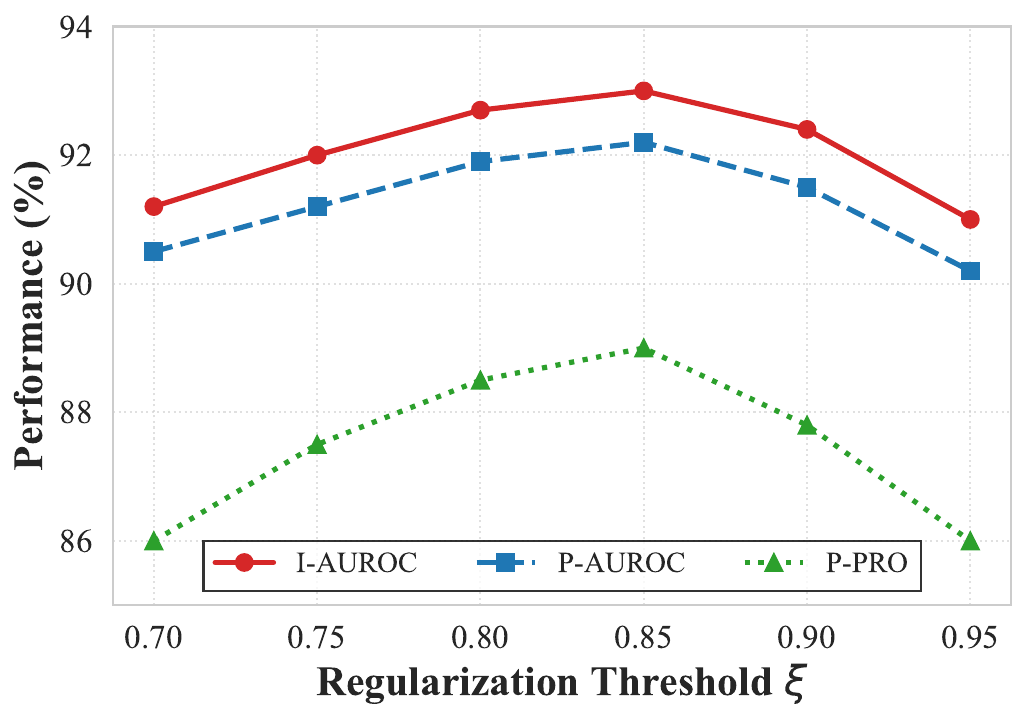}
        \caption{Analysis of the margin threshold $\xi$.}
        \label{fig:xi_analysis}
    \end{minipage}
    \vspace{-3mm} 
\end{figure}
Figures \ref{fig:gamma_analysis} and \ref{fig:xi_analysis} illustrate the sensitivity of the balance factor $\gamma$ and margin threshold $\xi$. Performance peaks at $\gamma=0.5$, which optimally balances defect detection with noise. $\xi=0.85$ effectively regularizes semantics, preventing drift while maintaining sufficient diversity to capture diverse anomalies.

\section{Conclusion}
\label{sec:conclusion}

In this paper, we presented SSVP, a novel framework that bridges the granularity gap in zero-shot industrial anomaly detection by establishing a deep synergy between CLIP's semantic generalization and DINOv3's structural discrimination. By combining Hierarchical Semantic-Visual Synergy (HSVS), Vision-Conditioned Prompt Generator (VCPG), and locally calibrated scoring (VTAM), our method transforms static semantic matching into an adaptive, visually grounded detection process. Extensive experiments across seven diverse benchmarks demonstrate that SSVP sets a new state-of-the-art, effectively handling both texture-centric and object-centric defects without target-domain supervision.
Despite these advancements, a primary limitation lies in the computational overhead incurred by the dual-backbone architecture and the iterative generative process, which restricts inference speed for real-time edge deployment. Future work will focus on two directions: (1) investigating knowledge distillation techniques to compress the synergistic knowledge into a lightweight unified network; (2) exploring test-time adaptation strategies to further refine the visual-semantic alignment on unlabelled target streams.




\bibliographystyle{plainnat}
\bibliography{cite}

\end{document}